\documentclass[sigconf]{acmart}

\usepackage{graphicx}
\usepackage{amsfonts}
\usepackage{subfigure}
\usepackage{subcaption}
\usepackage{multirow}
\usepackage{multicol}   
\usepackage{dashrule}   
\usepackage[para]{footmisc}
\usepackage{graphics}
\usepackage{soul}
\usepackage{float}
\usepackage{tcolorbox}
\usepackage{enumitem}
\usepackage{colortbl}
\definecolor{verylightgreen}{RGB}{245,255,245}

\AtBeginDocument{%
  }

\copyrightyear{2026}
\acmYear{2026}
\setcopyright{cc}
\setcctype{by}
\acmConference[WWW '26]{Proceedings of the ACM Web Conference 2026}{April 13--17, 2026}{Dubai, United Arab Emirates}
\acmBooktitle{Proceedings of the ACM Web Conference 2026 (WWW '26), April 13--17, 2026, Dubai, United Arab Emirates}
\acmPrice{}
\acmDOI{10.1145/3774904.3792931}
\acmISBN{979-8-4007-2307-0/2026/04}

\begin{document}

\newcommand{\Yu}[1]{{\color{orange}[Yu: #1]}}

\title{Multi-Agent Collaborative Filtering: \\ Orchestrating Users and Items for Agentic Recommendations}

\author{Yu Xia}
\affiliation{%
  \institution{University of California, San Diego}
  \city{La Jolla}
  \state{CA}
  \country{USA}
}
\email{yux078@ucsd.edu}

\author{Sungchul Kim}
\affiliation{%
  \institution{Adobe Research}
  \city{San Jose}
  \state{CA}
  \country{USA}
}
\email{sukim@adobe.com}

\author{Tong Yu}
\affiliation{%
  \institution{Adobe Research}
  \city{San Jose}
  \state{CA}
  \country{USA}
}
\email{tyu@adobe.com}

\author{Ryan A. Rossi}
\affiliation{%
  \institution{Adobe Research}
  \city{San Jose}
  \state{CA}
  \country{USA}
}
\email{ryrossi@adobe.com}

\author{Julian McAuley}
\affiliation{%
  \institution{University of California, San Diego}
  \city{La Jolla}
  \state{CA}
  \country{USA}
}
\email{jmcauley@ucsd.edu}

\renewcommand{\shortauthors}{Yu Xia, Sungchul Kim, Tong Yu, Ryan A. Rossi, and Julian McAuley}



\begin{abstract}
Agentic recommendations cast recommenders as large language model (LLM) agents that can plan, reason, use tools, and interact with users of varying preferences in web applications. 
However, most existing agentic recommender systems focus on generic single-agent plan-execute workflows or multi-agent task decomposition pipelines.
Without recommendation-oriented design, they often underuse the collaborative signals in the user–item interaction history, leading to unsatisfying recommendation results.
To address this, we propose the Multi-Agent Collaborative Filtering (MACF) framework for agentic recommendations, drawing an analogy between traditional collaborative filtering algorithms and LLM-based multi-agent collaboration.
Specifically, given a target user and query, we instantiate similar users and relevant items as LLM agents with unique profiles. 
Each agent is able to call retrieval tools, suggest candidate items, and interact with other agents.
Different from the static preference aggregation in traditional collaborative filtering,
MACF employs a central orchestrator agent to adaptively manage the collaboration between user and item agents via dynamic agent recruitment and personalized collaboration instruction.
Experimental results on datasets from three different domains show the advantages of our MACF framework compared to strong agentic recommendation baselines.
\end{abstract}

\ccsdesc[500]{Information systems~Recommender systems}

\keywords{Agentic Recommendation, LLM-based Multi-Agent System}



\maketitle

\section{Introduction}\label{sec:introduction}
Recommender systems are shifting from static predictors toward interactive agents. 
Recent agentic recommenders in web applications, such as online shopping assistants and movie finders, cast the system as a large language model (LLM) agent that can plan, reason, call tools, and interact with users \cite{huang2025towards,xia2025selection}.
These systems usually follow one of two patterns. 
The first uses a single LLM agent in a plan–and–execute loop \cite{xia2025sand,huang2025recommender} that reasons step-wise and calls retrieval or recommendation tools but still acts as a single decision-maker, making it hard to draw on diverse preference signals \cite{shen2025simultaneous}. 
The second organizes multiple LLM agents into task-decomposition pipelines with generic dialogue or planning roles \cite{wang2024macrec,fang2024multi,yu2025thought}. 
These multi-agent designs introduce role specialization and coordination,
but their roles are loosely tied to preference signal sources such as similar users or relevant items. 
Thus, while these approaches improve interactivity and tool use, they lack an effective way to bring collaborative signals into the agentic recommendation process.

As the foundation for many modern recommender systems, collaborative filtering (CF) has long shown that similar users and relevant items hold strong predictive value \cite{resnick1994grouplens,sarwar2001itemcf,xia2023user}.
Recent AgentCF \cite{zhang2024agentcf} extends this idea by modeling users and items as LLM agents that interact under logged user–item histories for behavior simulation.
Despite these advances, most LLM-based agentic recommenders still treat user–item histories only as context in prompts and do not structure or combine neighborhood evidence as CF does. 
As a result, they struggle to take advantage of the collaborative signals during recommendation, especially when similar users and correlated items provide complementary information.


\begin{figure*}[t!]
    \centering
    \includegraphics[width=1\textwidth]{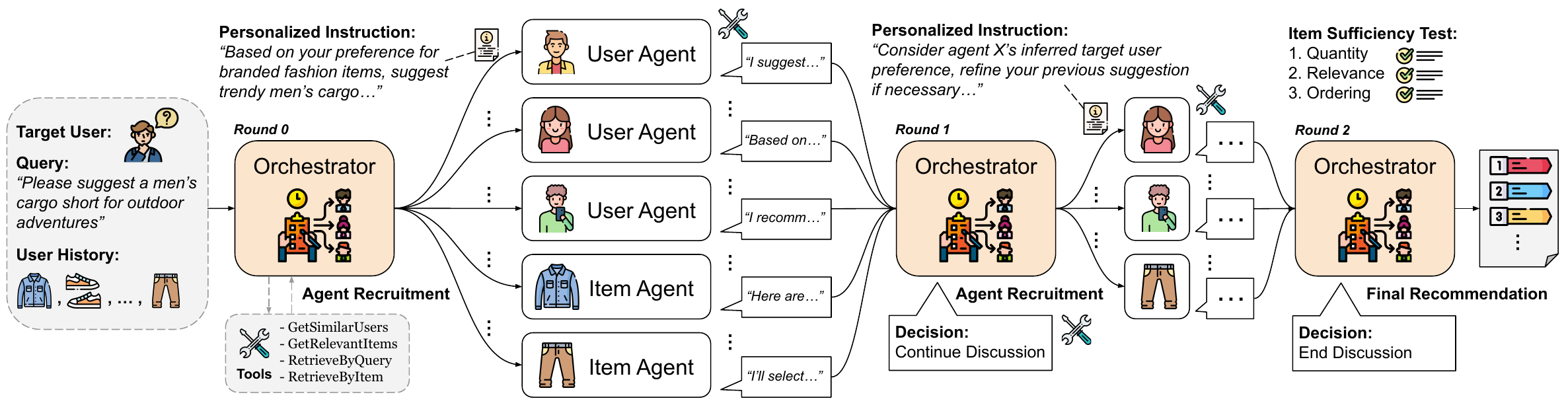}
    \caption{Overview of our MACF workflow illustrated with a product recommendation example  from Amazon Clothing.}
    \label{fig:overview}
\end{figure*}

To address this, we propose \textbf{M}ulti-\textbf{A}gent \textbf{C}ollaborative \textbf{F}iltering (MACF), which draws an analogy between traditional CF and LLM-based multi-agent collaboration.
Instead of using a single agent or generic multi-agent roles, MACF instantiates user agents and item agents that correspond to similar users and relevant items in the target user’s history. 
Each agent is initialized with a profile containing user information or item attributes, and can call retrieval tools, suggest candidate items, and exchange messages with other agents.
A central orchestrator is employed to manage a multi-round discussion between user and item agents. 
It dynamically selects which agents should participate in each round and issuing personalized instructions for each agent that shape how they contribute to the discussion.
This setup allows the system to draw on complementary evidence from neighborhoods on both the user and item side, refine candidates through agent communication, and surface agreements or conflicts that would be missed by fixed aggregation rules.
In this way, MACF leverages collaborative signals from user–item history in a structured way and produces recommendations with clear reasoning.
We evaluate MACF on three datasets from different domains, where it shows consistent improvements over strong agentic recommendation baselines, validating the advantages of our agentic collaborative filtering framework.
In summary, we make the following contributions:
\begin{itemize}[left=0pt]
\item We propose {MACF}, an agentic recommendation framework that draws an analogy between traditional collaborative filtering and LLM-based multi-agent collaboration.
\item We develop a dynamic orchestration mechanism that manages user and item agents across multiple rounds with personalized instructions, allowing MACF to aggregate collaborative signals in a structured and adaptive manner.
\item Experimental results demonstrate the advantage of our proposed MACF with consistent improvements over strong baselines on three datasets from different domains.
\end{itemize}

\section{Methodology}\label{sec:method}
In this section, we introduce our MACF framework as illustrated in Figure \ref{fig:overview}, which bridges traditional collaborative filtering and LLM-based multi-agent collaboration for agentic recommendation.

\subsection{Problem Definition}
Let $\mathcal{U}$ and $\mathcal{I}$ denote the sets of users and items.  
Each user $u\in\mathcal{U}$ has an interaction history 
$H_u = \langle i_1,\ldots,i_m \rangle$,
represented as an ordered list of items.  
Given a natural-language user query $q$, the task is to produce a ranked list 
$\hat{R}_{u,q}=\langle i_1,\ldots,i_K\rangle$ that reflects both the short-term user intent in $q$ 
and the long-term user preferences expressed in $H_u$.
We assume access to the following retrieval utilities:
\begin{itemize}[left=0pt]
    \item $\mathrm{GetSimilarUsers}(u,n)$ returns a set $\mathcal{N}_u$ of $n$ users  
whose preferences are most similar to the target user $u$;
    \item $\mathrm{GetRelevantItems}(u,q,n)$ returns a subset $\mathcal{H}_{u,q}$  
containing up to $n$ items from $H_u$ that are most relevant to the user query $q$; 
    \item $\mathrm{RetrieveByQuery}(q,k)$ retrieves $k$ items whose embeddings  
    are most similar to the user query $q$;
    \item $\mathrm{RetrieveByItem}(i,k)$ retrieves $k$ items whose embeddings  
    are most similar to a given item $i$.
\end{itemize}
These utilities are exposed to all MACF agents as callable tools $\mathcal{T}$.

\subsection{Agents Instantiation}
MACF instantiates two types of recommendation agents, user agents and item agents, that mirror the collaborative filtering structure, together with an orchestrator agent that coordinates their interaction for a given target user $u$, query $q$, and user history $H_u$.

\subsubsection*{\textbf{User Agents.}}
A {user agent} $a^{\text{user}}_v$ is instantiated for each selected similar user $v\in\mathcal{N}_u$. 
Each agent is initialized with a compact profile summarizing the user’s preference pattern and interaction history, along with the target user information.
From its own perspective, the user agent reasons about how the target user $u$ may align with itself, identifies which aspects of the target user’s preference are most informative for the query $q$, and prepares to suggest or critique items.
When needed, it calls retrieval tools in $\mathcal{T}$ to search for items that match the inferred preferences or a strong anchor item.

\subsubsection*{\textbf{Item Agents.}}
An {item agent} $a^{\text{item}}_j$ is instantiated for each selected query-relevant history item $j\in\mathcal{H}_{u,q}$ from target user history. 
Each agent is initialized with a compact profile of the item attributes and metadata. 
Given the target user information, the item agent reasons about why the target user $u$ previously interacted with item $j$ and how this evidence relates to the current user intent expressed in the query $q$.
It can optionally call retrieval tools in $\mathcal{T}$ to retrieve items closely related to itself or to broaden the search using the query when it provides more informative guidance.

\subsubsection*{\textbf{Orchestrator Agent.}}
The {orchestrator agent} $a^{\text{orc}}$ controls how user and item agents are instantiated and coordinated across a multi-round discussion.
Given the target user information, the orchestrator agent first calls tools $\mathcal{T}$ to select which similar users and relevant items to be instantiated as user agents $a^{\text{user}}$ and item agents $a^{\text{item}}$.
It then issues concise, personalized instructions for each agent that define their task in the current round.
Across rounds, the orchestrator continues to decide whether to continue the discussion, which agents remain active, how they are instructed, and how their suggestions and messages are aggregated into an evolving ranked list for the final recommendation.
We describe the detailed orchestration mechanism in Section~\ref{sec:macf}.

\subsection{Multi-Agent Collaborative Filtering}\label{sec:macf}
MACF runs a coordinated multi-agent discussion over rounds $t=0,1,\ldots,T_{\max}$. 
Each round produces agent messages and item candidates with rationales, together with a draft ranked list $\hat{R}_t$. 
The design mirrors collaborative filtering in language form: user agents bring preference evidence from similar users, item agents expand around history items, and the orchestrator adapts the mix to the target query $q$ through targeted coordination decisions.

\subsubsection*{\textbf{Dynamic Agent Recruitment}}
At the beginning of the discussion $t=0$, the orchestrator agent $a^{\text{orc}}$ calls the retrieval tools in $\mathcal{T}$ to select a set of similar users and query-relevant history items given target user $u$, query $q$, and user history $H_u$:
\[
\mathcal{N}_u = \mathrm{GetSimilarUsers}(u,n), \;
\mathcal{H}_{u,q} = \mathrm{GetRelevantItems}(u,q,n).
\]
It further filters these sets by relevance to the query $q$ and instantiates user agents $a^{\text{user}}$ and item agents $a^{\text{item}}$ from the selected elements. 
This step grounds MACF’s agent population directly in collaborative signals rather than in generic role templates.

\subsubsection*{\textbf{Personalized Collaboration Instruction}}
Once the agent set $\mathcal{A}=a^{\text{user}}\cup a^{\text{item}}$ is chosen, the orchestrator issues personalized instructions to each agent $a\in\mathcal{A}$. 
These instructions describe the agent’s task in the current round $t$, such as proposing new candidates, refining earlier suggestions, resolving conflicts, or examining low-relevance items in the current draft of ranked list $\hat{R}_t$. 
Instructions depend on each agent’s unique profile, target user information, and the state of the draft ranked list, which ensures that different agents contribute complementary forms of evidence.
For example, user agents highlight neighborhood preference patterns from $\mathcal{N}_u$, while item agents trace relevance paths from the user history $\mathcal{H}_{u}$.

\subsubsection*{\textbf{User–Item Collaboration Orchestration}}
For each discussion round $t \geq 1$, the orchestrator agent $a^{\text{orc}}$ begins by drafting a ranked list $\hat{R}_t$ of size $K$ from currently accumulated candidates and rationales from all agents. 
It then applies a sufficiency test on $\hat{R}_t$ that requires:  
(i) at least $K$ unique items with clear relevance to the query $q$ and the inferred preference;  
(ii) item-level alignment with the rationales provided by agents; and  
(iii) a clear ordering by relevance.  
If all conditions are met or $t = T_{\max}$, the discussion terminates and $\hat{R}_t$ becomes the final ranked list.

If the sufficiency test is not satisfied and $t < T_{\max}$, the discussion continues. 
The orchestrator agent $a^{\text{orc}}$ then selects a subset of agents $\mathcal{A}_t \subseteq \mathcal{A}$ whose perspectives are most useful for improving $\hat{R}_t$, and issues updated collaboration instructions for each agent tailored to the sources of remaining uncertainty, such as resolving ties and conflicts or replacing low-relevance items.
Given these instructions, user agents $a^{\text{user}}$ refine or contest items using preference evidence drawn from their similar-user viewpoint in $\mathcal{N}_u$ as well as optional additional tool calling.
Item agents likewise expand around strong anchors in $\mathcal{H}_{u}$ using $\mathrm{RetrieveByItem}(j,k)$ or adjust off-topic entries by invoking $\mathrm{RetrieveByQuery}(q,k)$ to find more query-aligned alternatives.  
Each agent also observes the full multi-agent discussion history and can use this shared context to support, critique, or counter earlier arguments, which allows MACF to combine the strengths of user-based CF and item-based CF in a single coordinated process through language-guided interaction.

\subsubsection*{\textbf{Final Recommendation}}
When the sufficiency test is satisfied or the round limit $T_{\max}$ is reached, the discussion terminates and the most recent draft list $\hat{R}_t$ becomes the final recommendation $\hat{R}_{u,q}$.  
This list is generated directly by the orchestrator agent $a^{\text{orc}}$, which consolidates all item candidates and rationales collected across rounds from all user and item agents and produces the final ordering through its drafting and re-ranking steps.  


\section{Experiments}\label{sec:experiments}

\begin{table}[!t]
	\centering
	\caption{Main results of all compared methods across datasets.}\vspace{-.5\baselineskip}
	\label{tab:main}
	\resizebox{.478\textwidth}{!}{
	\begin{tabular}{l|cc|cc|cc} 
	\toprule
	 & \multicolumn{2}{c|}{Amazon Clothing}& \multicolumn{2}{c|}{Amazon Beauty}&  \multicolumn{2}{c}{Amazon Music}\\
	Methods & H@10&N@10 & H@10&N@10 & H@10&N@10 \\
	\midrule
        BM25 & 0.0801& 0.2543& 0.2171& 0.4546& 0.1628& 0.3336\\
        BGE-M3 & 0.3103& 0.4356& 0.2694& 0.4889& 0.1187& 0.2947\\
        BGE-Reranker & 0.2970& 0.5031& 0.2980& 0.4946& 0.1763& 0.3516\\
        \midrule
    ItemCF & 0.3804 & 0.6258 & 0.3199 & 0.5141 & 0.3106 & 0.5105\\
    UserCF & 0.3687 & 0.5926 & 0.3037 & 0.4998 & 0.2988 & 0.4780\\
        \midrule
        ReAct & 0.4217 & 0.6912& 0.3921 & 0.6012 & 0.2323 & 0.4012\\
        Reflexion & 0.4218 &0.7002& 0.4001 &0.6203  & 0.3234 &0.4900\\
        InteRecAgent & 0.4207 & 0.7035& 0.3919 & 0.6136 & 0.2919& 0.4585\\
        MACRec & 0.4205& 0.6846& 0.3834& 0.5999& 0.2634& 0.4876 \\
        MACRS & 0.3871& 0.6441 & 0.3156& 0.5399& 0.2348& 0.4769\\
        TAIRA & 0.4447 & 0.7182 & 0.4655 & 0.7520  & 0.3519 & 0.5540  \\
    \midrule
        \textbf{MACF$_\text{item-based}$} & 0.4363 & \textbf{0.7963}  & 0.5027 & 0.8179  & 0.5088 & \underline{0.7614}  \\
        \textbf{MACF$_\text{user-based}$} & \underline{0.5134} & 0.7927  & \underline{0.5368} & \textbf{0.8244} & \underline{0.5134} & 0.7574  \\
        \textbf{MACF} & \textbf{0.5238} & \underline{0.7942} & \textbf{0.5421} & \underline{0.8215}  & \textbf{0.5450} & \textbf{0.7668} 
\\
 	\bottomrule
	\end{tabular}}
    \vspace{-1.\baselineskip}
\end{table}

\subsection{Experimental Setup}
\subsubsection*{\textbf{Datasets and Metrics}}
We evaluate our method on the three Amazon datasets: {Clothing}, {Beauty}, and {Music} \cite{mcauley2015image}.
We adopt the same user simulator-driven evaluation protocol as in \cite{yu2025thought} to measure the recommendation quality.  
\textbf{H@10} (HitRatio@10) measures how often relevant items appear in the top-$10$ positions of the recommendation list.    
\textbf{N@10} (NDCG@10) evaluates the ranking quality of top-$10$ items in the recommendation list.
We adopt the same data splits and natural language query sets as in \citet{yu2025thought}.

\subsubsection*{\textbf{Baselines and Variants}}
We compare MACF against three types of baselines. 
Retrieval-based recommenders include \textbf{BM25}, \textbf{BGE-M3}, and \textbf{BGE-Reranker} \cite{li2023making}. 
Traditional collaborative filtering baselines include \textbf{ItemCF} \cite{sarwar2001itemcf} and \textbf{UserCF} \cite{resnick1994grouplens} which are slightly adapted to consider user query via embedding-based reranking of CF results. 
We also compare recent single-agent and multi-agent agentic recommenders including \textbf{ReAct} \cite{yao2023react}, 
\textbf{Reflexion} \cite{shinn2023reflexion}, \textbf{InteRecAgent} \cite{huang2025recommender}, \textbf{MACRec} \cite{wang2024macrec}, \textbf{MACRS} \cite{fang2024multi}, and \textbf{TAIRA} \cite{yu2025thought}. 
For our method, besides \textbf{MACF} with user-item collaboration as described in Section \ref{sec:macf}, we also report the results of two variants: \textbf{MACF$_\text{user-based}$} instantiating only user agents for multi-agent collaboration and similarly \textbf{MACF$_\text{item-based}$} with only item agents.

\subsubsection*{\textbf{Implementation Details}}
We implement the orchestrator agent, user agents, and item agents as MCP services using asynchronous HTTP stream transport with all retrieval utilities in $\mathcal{T}$ provided as MCP tools. 
All agents share the backbone LLM of \texttt{gpt-4o} with temperature $0.3$. 
The input arguments for retrieval tools are defaulted to $n = 5$ and $k = 15$, but agents may adaptively override them when calling the tools.
All embeddings as required by tools are computed with BGE-M3 model \cite{li2023making} for fair comparison of all methods.
The maximum number of discussion rounds $T_{\max}$ is set to 5.  
The number of items in the targeted ranked list $K$ is set to 10.



\subsection{Results}
\subsubsection*{\textbf{Main Results}}
As presented in Table \ref{tab:main}, across all three domains, MACF delivers clear gains over retrieval, CF-based, and agentic baselines. 
On Clothing, MACF achieves the best hit rate while maintaining strong ranking quality. On Beauty, it again reaches the highest hit rate and stays competitive at the top for NDCG@10. On Music, MACF leads on both metrics, reflecting its ability to combine neighborhood signals and history-based reasoning across different domains.
The user-only and item-only variants also perform strongly. MACF$_\text{user-based}$ shows competitive hit rates across domains, while MACF$_\text{item-based}$ produces high NDCG@10 scores, especially on Clothing and Beauty. 
These results highlight that grounding agents in collaborative filtering signals and guiding their interaction through the orchestrator consistently improves relevance and ranking accuracy.

\subsubsection*{\textbf{Ablation Studies}}
We report in Table~\ref{tab:ablation} performances of MACF after removing each key module.  
From the results, we observe that disabling personalized collaboration instruction (PCI) or dynamic agent recruitment (DAR) reduces both HR@10 and NDCG@10 across datasets, verifying the value of structured agent selection and targeted guidance in our orchestrator agent design.  
Turning off adaptive tool use (ATU), where each agent only call tools once at the beginning, also leads significant degradation, especially on Music, where relevant anchors vary more across user histories. 
These results further confirm the validity of our agent designs and dynamic orchestration mechanism for MACF.

\begin{table}[!t]
	\centering
	\caption{Ablation studies of different modules in MACF}
	\label{tab:ablation}\vspace{-.25\baselineskip}
	\resizebox{0.45\textwidth}{!}{
	\begin{tabular}{l|cc|cc|cc} 
	\toprule 
	 & \multicolumn{2}{c|}{Amazon Clothing} & \multicolumn{2}{c|}{Amazon Beauty} &  \multicolumn{2}{c}{Amazon Music}\\
  Variants & H@10& N@10 & H@10& N@10 & H@10& N@10 \\
  \midrule
    \textbf{MACF} & \textbf{0.5238} & \textbf{0.7942} & \textbf{0.5421} & \textbf{0.8215} & \textbf{0.5450} & \textbf{0.7668} \\
\midrule
    $w/o$ PCI & \underline{0.5165} & \underline{0.7902} & \underline{0.5147} & 0.7836 & 0.5372 & \underline{0.7417} \\
    $w/o$ DAR & 0.5029 & 0.7462 & 0.5053 & 0.7753 & \underline{0.5448} & 0.7129 \\
    $w/o$ ATU & 0.4300 & 0.7516 & 0.4890 & \underline{0.7882} & 0.4884 & 0.6682\\
    \bottomrule
	\end{tabular}} 
\end{table}

\begin{figure}[t!]
    \centering
    \includegraphics[width=0.478\textwidth]{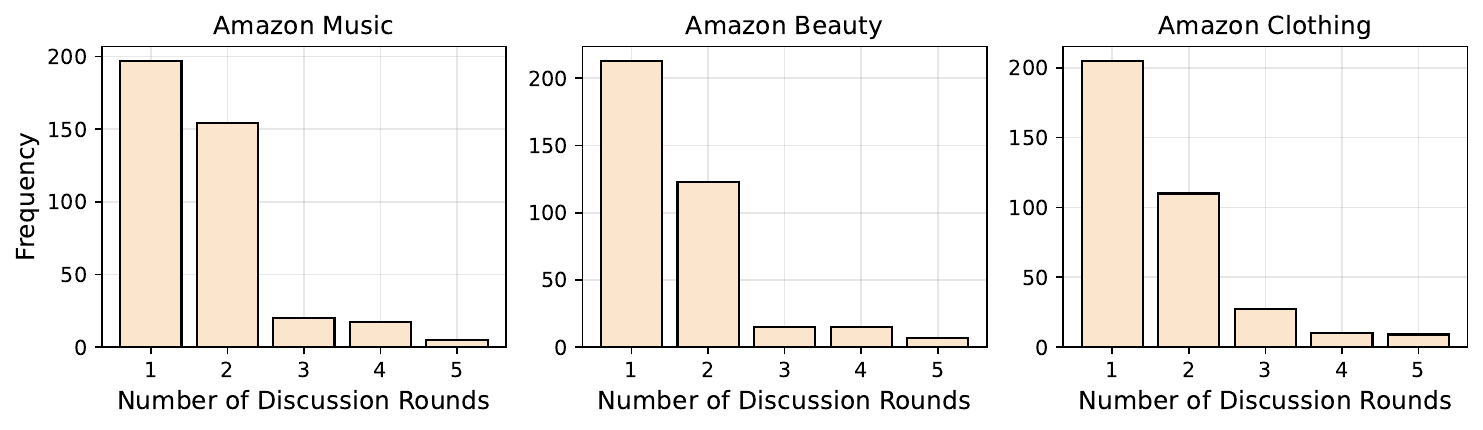}
    \vspace{-1.25\baselineskip}
    \caption{Distributions of MACF discussion rounds.}
    \label{fig:round_dist}
\end{figure}

\subsubsection*{\textbf{Additional Analysis}}
To further understand the behavior of MACF during multi-agent discussion, we show in Figure~\ref{fig:round_dist} the distribution of discussion rounds across three datasets.
We observe that most queries finish before the third round, with very few cases reaching the set maximum of five rounds. This indicates that our designed sufficiency test in collaboration orchestration is effective at identifying when the ranked list is ready to be recommended to the user, preventing unnecessary interaction and reducing extra cost.

\section{Conclusion}\label{sec:conclusion}
In this paper, we present Multi-Agent Collaborative Filtering (MACF), an agentic recommendation framework that grounds LLM-based multi-agent collaboration in collaborative filtering signals. 
By instantiating similar users and relevant items as LLM agents and coordinating them with a central orchestrator, MACF turns user–item evidence into an interactive, query-aware multi-agent discussion that yields stronger recommendations. Experiments on three datasets show consistent gains over traditional recommenders and strong agentic recommendation baselines, highlighting the potential of agentic modeling for future recommender systems.

\section*{Acknowledgement}
This work is partially supported by NSF IIS-2432486.

\bibliographystyle{ACM-Reference-Format}
\balance
\bibliography{ref}



\end{document}